\pdfoutput=1

\documentclass[11pt]{article}

\usepackage[final]{acl}

\usepackage{times}
\usepackage{latexsym}

\usepackage[T1]{fontenc}

\usepackage[utf8]{inputenc}

\usepackage{microtype}

\usepackage{inconsolata}

\usepackage{booktabs}

\usepackage{graphicx}
\usepackage{amsmath}
\usepackage{boldline}
\usepackage{enumitem}
\usepackage{multirow}

\title{DHP Benchmark: Are LLMs Good NLG Evaluators?}

\author{
 \textbf{Yicheng Wang\textsuperscript{1$\dagger$}},
 \textbf{Jiayi Yuan\textsuperscript{2$\dagger$}},
 \textbf{Yu-Neng Chuang\textsuperscript{2}},
 \textbf{Zhuoer Wang\textsuperscript{1}},
 \textbf{Yingchi Liu\textsuperscript{3}},
 \\
 \textbf{Mark Cusick\textsuperscript{3}},
 \textbf{Param Kulkarni\textsuperscript{3}},
 \textbf{Zhengping Ji \textsuperscript{3}},
 \textbf{Yasser Ibrahim\textsuperscript{3}},
 \textbf{Xia Hu\textsuperscript{2}},
\\
\\
 \textsuperscript{1}Texas A\&M University,
 \textsuperscript{2}Rice University,
 \textsuperscript{3}Axon Enterprise, Inc.
\\
}

\begin{document}
\maketitle
\renewcommand{\thefootnote}{\fnsymbol{footnote}}
\footnotetext[2]{Equal Contribution.}

\begin{abstract}
Large Language Models (LLMs) are increasingly serving as evaluators in Natural Language Generation (NLG) tasks; this is often referred to as ``LLM-as-a-judge'' paradigm. However, the capabilities of LLMs in evaluating NLG quality remain underexplored. Current studies depend on human assessments and simple metrics that fail to capture the discernment of LLMs across diverse NLG tasks. To address this gap, we propose the Discernment of Hierarchical Perturbation (DHP) benchmarking framework, which provides quantitative discernment scores for LLMs. This framework leverages hierarchically perturbed text data and statistical tests to systematically measure the NLG evaluation capabilities of LLMs. We re-established six evaluation datasets for this benchmark, covering four NLG tasks: Summarization, Story Completion, Question Answering, and Translation. Our comprehensive benchmarking of five major LLM families provides critical insight into their strengths and limitations as NLG evaluators. Our dataset is available at \url{https://huggingface.co/datasets/YCWANGVINCE/DHP_Benchmark}.
\end{abstract}

\section{Introduction}
Large Language Models (LLMs) play a crucial role in the field of Natural Language Generation (NLG), advanced wide real-world applications including education~\cite{latif2023knowledge}, healthcare~\cite{yuan2023large}, business~\cite{teubner2023welcome}, etc. The strong capabilities of LLMs allow them not only to serve as text generators but also increasingly as powerful evaluators of text quality~\cite{chiang2023can, liu2023g, li2024leveraging}. Their role as evaluators is crucial for advancements in various applications, such as summarization, story completion, question answering, and translation~\cite{li2024leveraging, wang2023chatgpt, chuang2024spec}. LLMs are expected to serve as NLG evaluators, providing reasonable quality scores based on different quality metrics with specially designed evaluation prompts.

Despite the growing performance of LLMs in evaluation tasks, a significant gap remains in fully comprehending their capabilities in evaluating NLG quality. The question, \textit{\textbf{Are LLMs good NLG evaluators?}} remains challenging for two main reasons illustrated in Figure~\ref{fig1}: 

\begin{figure*}[ht!]
    \centering
    \includegraphics[width=0.95\textwidth]{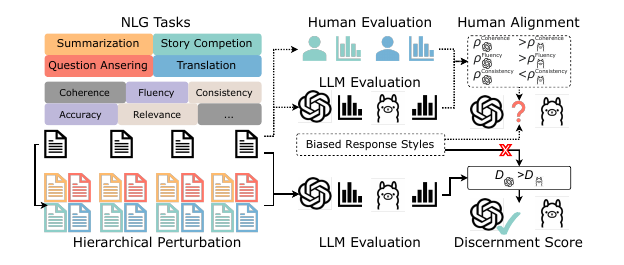}
    \caption{Challenges in Assessing LLMs as NLG Evaluators: Biased Response Styles and Multiple Evaluation Metrics. Our DHP Framework employs hierarchical perturbation and statistical tests to address these challenges, offering quantitative discernment scores for effective comparison.}
    \label{fig1}
\end{figure*}

\noindent(1) Lack of Clear and Unbiased Measurement: There is no clear measurement for the capability of LLM evaluators. Existing methods rely on aligning with human scores~\cite{chiang2023can, liu2023g}, but these scores themselves are subject to biased response styles~\cite{schoch2020problem}. 

\noindent(2) Multiple Evaluation Metrics: Evaluating NLG quality requires considering multiple metrics. For example, in summarization tasks, metrics such as coherence, consistency, and fluency are essential considerations~\cite{liu2023g, fabbri2021summeval, gabriel-etal-2021-go}. However, LLMs might struggle with correlations between these metrics~\cite{hu2024llm}, potentially leading to misinterpretation and incorrect scoring, which makes it difficult to assess their effectiveness as evaluators.

To address these challenges, we introduce a novel \textbf{DHP benchmarking framework} -- \textbf{D}iscernment of \textbf{H}ierarchical \textbf{P}erturbation -- for quantitatively measuring the evaluation capabilities of LLMs. We propose the concept of discernment scores, systematically derived from hierarchically perturbed text data and statistical tests. Reference data is perturbed using multiple hierarchical methods, and differences in LLM evaluation scores are analyzed using the Wilcoxon Signed-Rank Test~\cite{wilcoxon1945individual}. To obtain more reliable overall evaluation results, harmonic mean $p$-values and expert-assigned weights are applied to integrate multiple metrics. The final $p$-value is then converted into a \textbf{Discernment Score}, providing a quantitative measure of the NLG evaluation capabilities of LLMs. This approach enables a more rigorous and comprehensive assessment of LLM performance, independent of the specific response styles exhibited by the models.

This study re-establishes six evaluation datasets across four key NLG tasks: Summarization, Story Completion, Question Answering, and Translation. Each dataset undergoes hierarchical perturbation and is utilized to challenge the evaluative capabilities of LLMs in distinct ways, providing a robust foundation for benchmarking. The datasets include a range of text perturbation methods, from minor character-level problems to significant sentence-level alterations, enabling a thorough examination of the potential discernment limits of LLMs.

Our comprehensive benchmarking, based on newly defined quantitative discernment scores, is conducted across five major LLM series. This methodology uncovers critical insights into their effectiveness as NLG evaluators and provides a detailed understanding of their performance. This benchmark reveals important trends and patterns in the LLM evaluator capacities, highlighting areas of strength as well as potential shortcomings.

The DHP benchmark aims to fill existing gaps by offering a quantitative framework for assessing LLMs' evaluation capabilities and emphasizing the necessity of considering multiple metrics for accurate and reliable evaluations. We summarize our contributions as follows.

\begin{itemize}[leftmargin=*, nosep]
    \item Develop the DHP benchmarking framework, introducing quantitative discernment scores for LLMs as NLG evaluators based on hierarchical perturbation.

    \item Re-establish six evaluation datasets across four NLG tasks to evaluate the discernment of LLM evaluators.

    \item Benchmark five LLM series to analyze their performance and effectiveness in NLG evaluation.
    
\end{itemize}

\begin{figure*}[t]
    \centering
    \includegraphics[width=0.95\textwidth]{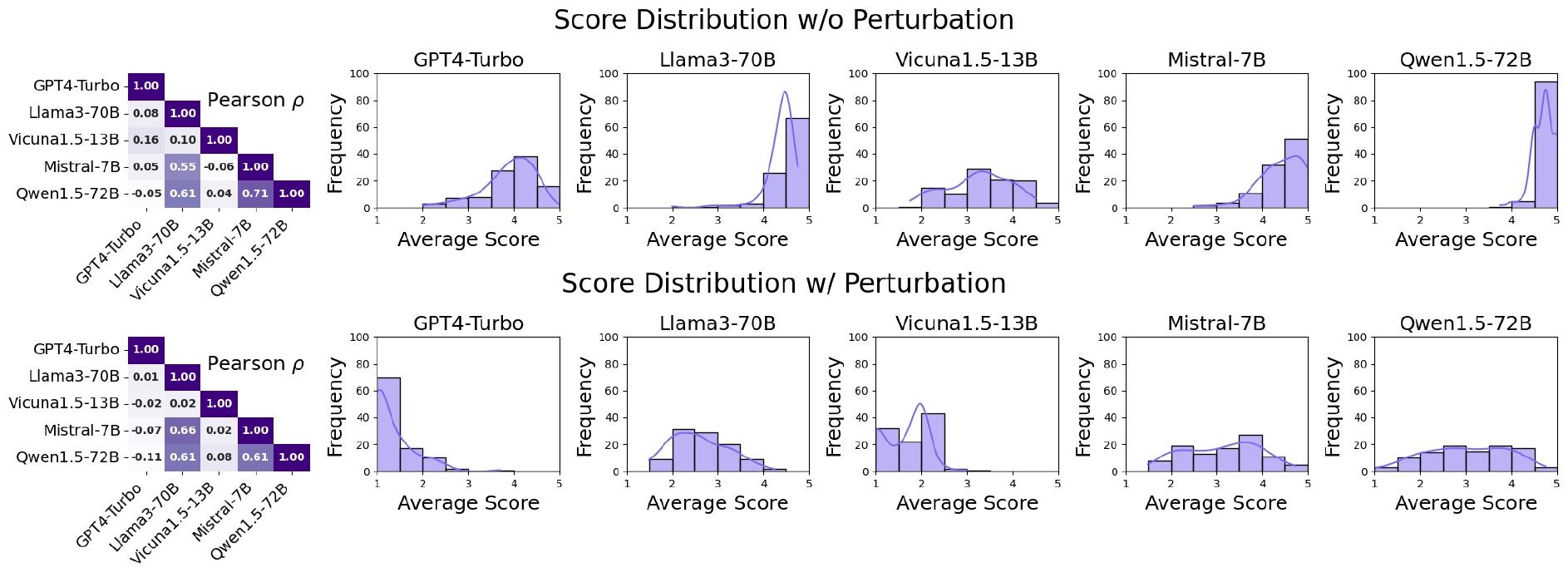}
    \caption{Response styles of five LLMs evaluated using the SummEval dataset~\cite{fabbri2021summeval}.}
    \label{fig2}
\end{figure*}

\section{Challenge: Biased Response Styles}
Previous studies focus on the alignment between human and LLM evaluators, using correlation metrics to gauge the LLMs' performance in NLG evaluation tasks~\cite{liu2023g, chiang2023can}. However, these studies often overlook an important variable of evaluators: \textbf{Response Styles} which refer to a respondent's consistent manner of answering survey questions, regardless of the content~\cite{van2013response}. Despite similar levels of professionalism, annotators may assign different scores to the same questionnaire due to differences in age, gender, personality, cultural background, and ethnic group~\cite{van2013response,hui1989effects,kieruj2013response}. Similarly, LLMs, trained on diverse datasets, may also exhibit biases in their responses~\cite{salecha2024large}. This discrepancy casts doubt on the previous methods used to compare human and LLM scores. Since quality-based scoring often relies heavily on a few experts' annotations, the final alignment scores tend to favor models that share similar response styles with these specific experts.

We illustrate this with an example of the response styles of five LLMs tasked with annotating quality scores for human reference data from the SummEval dataset~\cite{fabbri2021summeval}. We averaged the scores across four metrics for each data point and plotted both the Pearson correlation coefficient ($\rho$) and the average score distributions of the five models. After perturbing the original data by replacing some named entities with fictional ones in the summaries (Fictional Named Entities in Table~\ref{table1}), we repeated the quality evaluation. As shown in Figure~\ref{fig2}, all models detected the changes and adjusted their scores accordingly, though their scoring distributions varied significantly. For instance, Llama3~\cite{meta2024llama3}, Mistral~\cite{jiang2023mistral}, and Qwen~\cite{qwen} models assign higher scores to the original data and moderate scores to the perturbed data. In contrast, GPT4-Turbo~\cite{openai2023gpt4turbo} and Vicuna~\cite{vicuna2023} models tend to give moderate scores to the original data and much lower scores to the perturbed data. The variance in the response distributions indicates the presence of bias that can significantly affect alignment ($\rho$), illustrating that alignment is not a direct or credible metric for assessing the ability of LLMs as NLG evaluators. It is crucial to develop a new metric and measurement for evaluation that is not influenced by the evaluators' biased response styles, ensuring a more accurate and fair assessment of LLM capabilities.

\begin{figure*}[t]
    \centering
    \includegraphics[width=0.95\textwidth]{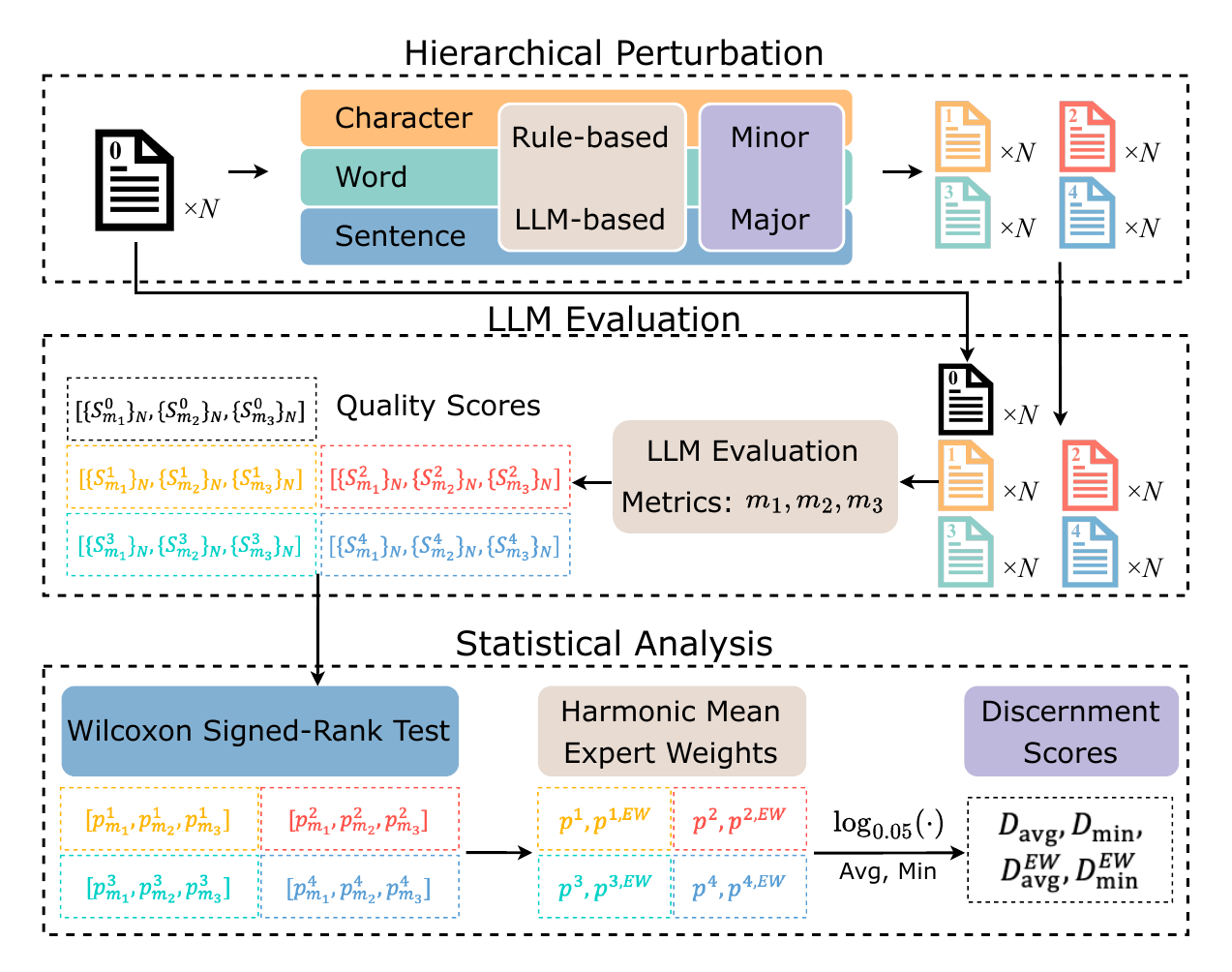}
    \caption{The DHP framework for each NLG task. It includes three steps: (1) Hierarchical Perturbation, (2) LLM Evaluation, and (3) Statistical Analysis. This figure demonstrates the framework with four perturbation types ($P=4$) and three evaluation metrics ($M=3$).}
    \label{fig3}
\end{figure*}

\section{DHP Benchmarking Framework}

We propose our DHP framework: Discernment of Hierarchical Perturbation. Previous studies overlook the essence of NLG evaluation, i.e., the content-oriented scoring~\cite{novikova2018rankme}. In other words, content that is accurate, fluent, and consistent should receive higher scores than content that is inaccurate, disfluent, and inconsistent. Qualified annotators should be able to recognize inappropriate content without additional references and then assign scores, even though the absolute scores may still reflect their biased response styles. The fundamental principle of our assessment is that a qualified LLM evaluator should be able to independently identify issues in perturbed data (which contains some quality issues) and assign relatively lower scores compared to the original reference data during two separate evaluations. This approach does not rely on human scores, thus eliminating the influence of human response styles.

The overall framework is shown in Figure~\ref{fig3}. First, for a specific NLG task, we employ a hierarchical perturbation pipeline to transform high-quality reference data into various forms of lower-quality data. Subsequently, an LLM evaluates both the original and perturbed texts, respectively, using predefined metrics, generating several sets of rating scores. We then conduct a statistical analysis of these scores. For each pair of scores, original and perturbed, we apply the Wilcoxon Signed-Rank Test to determine the differences in their distributions, achieving this with a confidence level expressed as a $p$-value. This test specifically assesses differences in pairwise scores without focusing on absolute values, thereby minimizing the impact of models' response styles. Following this, we combine the $p$-values from different metrics, incorporating Expert Weights ($EW$) to tailor the aggregated $p$-values to the specific metrics of the corresponding perturbation methods. These combined $p$-values are then transformed into discernment scores, which serve as a direct measure for assessing and comparing the NLG evaluation capabilities of LLMs for this particular task.

\subsection{Step 1: Hierarchical Perturbation}

To generate data that have quality issues across various levels, formats, and evaluation difficulties, we propose a hierarchical perturbation approach. In contrast to the plain perturbations~\cite{evaleval}, our approach encompasses three levels of perturbation content: character, word, and sentence levels; two methods of perturbation: rule-based and LLM-based; and two degrees of perturbation: minor and major as illustrated in Figure~\ref{fig3}. 

First, at the character level, we alter some characters or letters in the given $N$ original texts independently. At the word and sentence levels, we degrade the text by processing entire words or sentences, respectively. For NLG tasks involving very short texts, sentence-level perturbation is considered optional. For each level of perturbation, we choose either a rule-based or an LLM-based method, enhancing the diversity of the perturbation's content and format. Additionally, if the text data is sufficiently long for more perturbation, we implement two degrees of perturbation--minor and major--for each method. These different degrees of perturbation are the difficulty that LLMs face in detecting issues within the text. The detailed perturbation methods for each task are shown in Table~\ref{table1}.

With this approach, we generate multiple sets of perturbed data, with each set designed to highlight a specific quality issue tied to a distinct type of perturbation method. Competent LLM evaluators should accurately detect these issues and assign correspondingly lower scores to the perturbed data.

\begin{table*}[t]
\small
\centering
\caption{The quality metrics and perturbation methods for the four NLG tasks. \textcolor{RedOrange}{\textbf{C}}: Character Level. \textcolor{Green}{\textbf{W}}: Word Level. \textcolor{Cyan}{\textbf{S}}: Sentence Level. \textcolor{Tan}{\textbf{(R)}}: Rule-based Perturbation. \textcolor{Tan}{\textbf{(L)}}: LLM-based Perturbation. \textcolor{Mulberry}{\textbf{(M)}}: Major and Minor Perturbations for each method.}
\scalebox{0.9}{
\begin{tabular}{lll}
\toprule
\textbf{Task}                                                          & \textbf{Metrics}                                                                                     & \textbf{Perturbations}                                                                                                                                                                                            \\ \midrule
Summarization                                                 & \begin{tabular}[c]{@{}l@{}}Coherence \\ Consistency \\ Fluency \\ Relevance\end{tabular} & \begin{tabular}[c]{@{}l@{}}\textcolor{RedOrange}{\textbf{C}} \textcolor{Mulberry}{\textbf{(M)}}: Random Deletions \textcolor{Tan}{\textbf{(R)}}, Random Typos \textcolor{Tan}{\textbf{(R)}}\\ \textcolor{Green}{\textbf{W}} \textcolor{Mulberry}{\textbf{(M)}}: Fictional Named Entities \textcolor{Tan}{\textbf{(L)}}, Grammatical Errors \textcolor{Tan}{\textbf{(L)}}\\ \textcolor{Cyan}{\textbf{S}} \textcolor{Mulberry}{\textbf{(M)}}: Reordering \textcolor{Tan}{\textbf{(R)}}, Rewriting and Insertion \textcolor{Tan}{\textbf{(L)}}\end{tabular} \\ \midrule
\begin{tabular}[c]{@{}l@{}}Story \\ Completion\end{tabular}   & \begin{tabular}[c]{@{}l@{}}Coherence \\ Consistency\\ Fluency\end{tabular}                & \begin{tabular}[c]{@{}l@{}}\textcolor{RedOrange}{\textbf{C}}: Random Deletions \textcolor{Tan}{\textbf{(R)}}, Random Typos \textcolor{Tan}{\textbf{(R)}}\\ \textcolor{Green}{\textbf{W}}: Fictional Named Entities \textcolor{Tan}{\textbf{(L)}}, Grammatical Errors \textcolor{Tan}{\textbf{(L)}}\\ \textcolor{Cyan}{\textbf{S}}: Random Ending Sentence \textcolor{Tan}{\textbf{(R)}}, Wrong Ending Sentence \textcolor{Tan}{\textbf{(R)}}\end{tabular}   \\ \midrule
\begin{tabular}[c]{@{}l@{}}Question \\ Answering\end{tabular} & Answer Quality                                                                              & \begin{tabular}[c]{@{}l@{}}\textcolor{RedOrange}{\textbf{C}} \textcolor{Mulberry}{\textbf{(M)}}: Random Deletions \textcolor{Tan}{\textbf{(R)}}, Random Typos \textcolor{Tan}{\textbf{(R)}}\\ \textcolor{Green}{\textbf{W}} \textcolor{Mulberry}{\textbf{(M)}}: Fictional Named Entities \textcolor{Tan}{\textbf{(L)}}, Grammatical Errors \textcolor{Tan}{\textbf{(L)}}\\ \textcolor{Cyan}{\textbf{S}}: Random Answer \textcolor{Tan}{\textbf{(R)}}\end{tabular}                               \\ \midrule
Translation                                                   & \begin{tabular}[c]{@{}l@{}}Accuracy \\ Fluency\end{tabular}                                & \begin{tabular}[c]{@{}l@{}}\textcolor{RedOrange}{\textbf{C}} \textcolor{Mulberry}{\textbf{(M)}}: Random Deletions \textcolor{Tan}{\textbf{(R)}}, Random Typos \textcolor{Tan}{\textbf{(R)}}\\ \textcolor{Green}{\textbf{W}} \textcolor{Mulberry}{\textbf{(M)}}: Random Deletions \textcolor{Tan}{\textbf{(R)}}, Fictional Named Entities \textcolor{Tan}{\textbf{(L)}}, Grammatical Errors \textcolor{Tan}{\textbf{(L)}}\end{tabular}                 \\ \bottomrule

\end{tabular}
}
\label{table1}

\end{table*}

\subsection{Step 2: LLM evaluation}

Following the evaluation method outlined in G-Eval~\cite{liu2023g}, we also utilize the automatic chain-of-thought approach (Auto-CoT)~\cite{zhang2022automatic} to design evaluation prompts for different datasets and evaluation metrics. These prompts are sent to LLMs to assess both the original data and the perturbed, low-quality data. It's important to note that all perturbed data are evaluated independently, without their original references, to accurately test the models' capabilities in identifying specific quality issues.

After conducting the LLM evaluation on $N$ datapoints, we obtain several sets of absolute evaluation scores shown in Figure~\ref{fig3}:

\begin{align*}
    &[ \{S^0_{m_1}\}, \{S^0_{m_2}\}, \dots, \{S^0_{m_M}\} ], \\
    &[ \{S^1_{m_1}\}, \{S^1_{m_2}\},  \dots, \{S^1_{m_M}\} ], \cdots, \\
    &[ \{S^P_{m_1}\}, \{S^P_{m_2}\}, \dots, \{S^P_{m_M}\} ], \\
\end{align*}

where each $\{S\}$ is a set of $N$ evaluation scores. The superscripts $0, 1, \ldots, P$ on $S$ represent the original data ($0$) and the $P$ types of perturbed data ($1, \dots, P$), respectively. The subscripts $m_1, \ldots, m_M$ represent the $M$ different metrics used in the dataset. For instance, in the SummEval dataset~\cite{fabbri2021summeval}, there are four evaluation metrics, namely: coherence, consistency, fluency, and relevance.

\subsection{Step 3: Statistical Analysis}

As illustrated in Figure~\ref{fig3}, we conduct a chain of statistical analyses to derive the final discernment scores for LLM evaluators. This process includes the Wilcoxon Signed-Rank Test, Harmonic Mean $p$-value and Expert Weights, and the final calculation of discernment scores. 

\subsubsection{Wilcoxon Signed-Rank Test}

The Wilcoxon Signed-Rank Test (W-Test)~\cite{wilcoxon1945individual} is a non-parametric hypothesis test that compares two dependent samples to assess whether their population mean ranks differ significantly. We apply the W-Test to evaluate whether there is a significant difference in the score distributions between the original data and a given type of perturbed data:

\begin{equation*}
    p^i_{m_j} \sim z^i_{m_j} = \text{W-Test}(\{S^0_{m_j}\}, \{S^i_{m_j}\}).
\end{equation*}

In our analysis, we adopt a one-sided alternative hypothesis. The resulting $p$-value indicates the confidence level at which we can reject the null hypothesis -- that $\{S^0_{m_j}\}$ and $\{S^i_{m_j}\}$ have the same distribution -- and accept the alternative hypothesis -- that $ \{S^0_{m_j}\}$ has a greater distribution than $\{S^i_{m_j}\}$. We consider a difference to be statistically significant if $p^i_{m_j} < 0.05$. A lower $p$-value represents a more significant score difference between the original data and perturbed data. In total, we can get $P$ sets of $p$-values for the $M$ metrics, as shown in Figure~\ref{fig3}:

\begin{equation*}
    [p^1_{m_1}, p^1_{m_2}, \dots, p^1_{m_M}],
    \cdots,
    [p^P_{m_1}, p^P_{m_2}, \dots, p^P_{m_M}].
\end{equation*}

Because the W-Test does not assume any specific distribution for the scores and does not focus on their absolute values, the resulting $p$-values solely reflect whether the LLMs are able to detect the quality issues and assign lower scores to the perturbed data compared to the original data. Consequently, this testing approach inherently avoids the influence of response styles, instead focusing on the relative quality assessment. Meanwhile, the $p$-values provide a quantitative evaluation measure to the score difference, i.e., the capability of evaluators to discern low-quality data.

\subsubsection{Harmonic Mean p-value and Expert Weights}

Given that an evaluation task may involve multiple $M$ evaluation metrics, resulting in multiple $p$-values $[p^i_{m_1}, p^i_{m_2}, \dots, p^i_{m_M}]$ for a single perturbed set, it is crucial to derive a combined $p$-value to measure the overall confidence level. We employ the Harmonic Mean $p$-value (HMP) method~\cite{wilson2019harmonic} without or with the Expert Weights ($EW$) presented in Figure~\ref{fig3}:

\begin{equation*}
    p^{i} = \frac{1}{\sum^M_{j=1} \frac{1}{p^i_{m_j}}}, \quad p^{i,EW} = \frac{1}{\sum^M_{j=1} \frac{EW^i_{m_j}}{p^i_{m_j}}}.
\end{equation*}

\begin{figure*}[!t]
    \centering
    \includegraphics[width=0.95\textwidth]{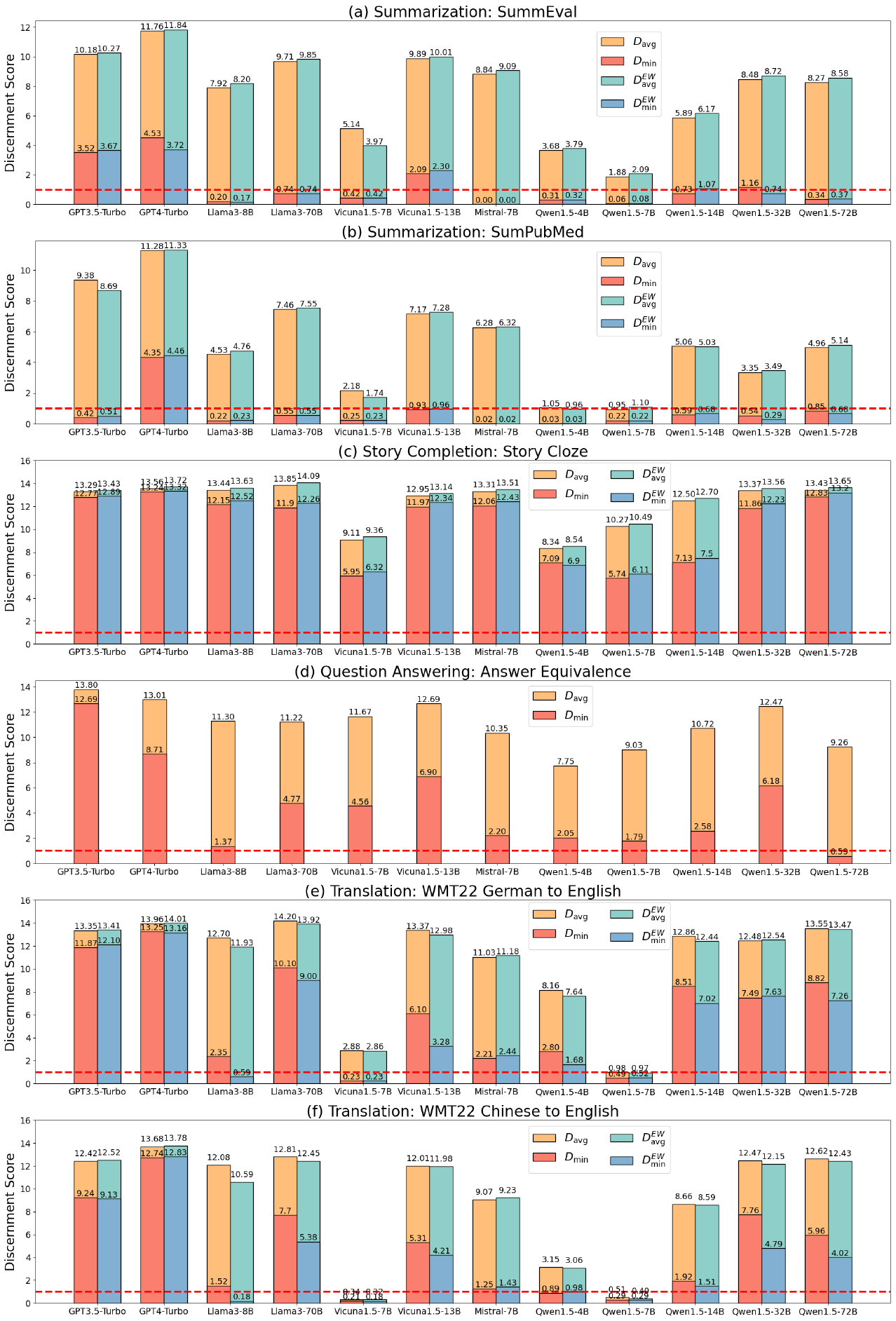}

    \caption{The DHP benchmarking results across four NLG tasks. Notably, in (d) for the Question Answering task, $D$ and $D^{EW}$ are identical because this task utilizes only one evaluation metric. The red lines on the charts represent $D$ or $D^{EW}=1$, which indicates the threshold for statistical significance in discernment scores.}
    \label{fig4}

\end{figure*}

There are two main reasons for using the HMP method: (1) The $p$-values are dependent as they are derived from the same dataset but differ based on potentially correlated metrics. The HMP method accommodates this dependency~\cite{wilson2019harmonic, vovk2020combining}. (2) The harmonic mean emphasizes the effect of smaller numbers, meaning that even if the LLMs identify and appropriately score a problem in just one metric, the combined $p$-value is still apparently small enough. However, a limitation of the simple HMP is that it does not indicate whether the LLM evaluators correctly identify the specific problems related to the corresponding metrics. For example, in the SummEval~\cite{fabbri2021summeval} dataset, if a perturbation targets the ``fluency'' metric but the LLM evaluator incorrectly assigns lower scores to ``relevance'', the Harmonic Mean $p$-value method might still produce a low combined $p$-value. This outcome may not accurately reflect the evaluator's ability to identify the specific issue.

To address this, we introduce HMP with Expert Weights ($EW$). We conduct a survey involving $10$ NLP experts who are presented with the specific NLG evaluation tasks and metric definitions. They are asked to identify which metric should be most impacted by different quality problems corresponding to the perturbation methods. These preferences are then aggregated to construct $EW$. For instance, a particular quality issue get votes for ``coherence'', ``consistency'', and ``fluency'' are $4, 1,$ and $5$, respectively, the $EW$ for the corresponding perturbation would be $[0.4, 0.1, 0.5]$. The $EW$ makes the combination more targeting those $p$-values that are highly influenced by the perturbation. This weighting makes the $p$-value combination more targeted, focusing on those metrics most influenced by the perturbation. Consequently, the weighted combined $p$-values offer a more precise measure of the LLM evaluators' ability to not only detect issues but also correctly assign lower scores to the impacted metrics.

\subsubsection{Discernment Scores of LLM Evaluators}

To facilitate comparisons, we transform these combined $p$-values into positive scores, which we define as discernment scores for a specific perturbation $i$ in Figure~\ref{fig3}:

\begin{equation*}
    D^i = \log_{0.05}(p^i), \quad D^{i,EW} = \log_{0.05}(p^{i,EW}).
\end{equation*}

Here, $D^i$ and $D^{i,EW}$ are positive values and the higher the better. A value of 1 for $D^i$ and $D^{i, EW}$ is a threshold corresponding to a $p$-value of 0.05, indicating statistical significance. If $D^i$ or $D^{i, EW}$ is less than 1, it means that the LLM evaluators do not assign significantly lower scores to the perturbed data compared to the original data, suggesting a lack of discernment for specific quality issues during the NLG evaluation.

To observe the comprehensive capability and worst-case performance of the LLMs, we calculate both the average and minimum of $D^i$ and $D^{i,EW}$ across all perturbation methods $i=1,\dots,P$. This results in overall LLM discernment scores $D_{\text{avg}}$, $D_{\text{min}}$, $D^{EW}_{\text{avg}}$, and $D^{EW}_{\text{min}}$. Note that the average discernment scores are calculated using a weighted average across the perturbation levels (character, word, and sentence levels) mentioned previously. We assign equal weights to perturbations within the same level and make sure that the sum of the weights is the same for each level. 
This weighting approach ensures that each level of perturbation contributes equally to the final scores.

These discernment scores allow us to explicitly evaluate and compare the capabilities of LLMs as evaluators on specific NLG tasks, thereby establishing comprehensive benchmarks for LLMs. Higher average discernment scores ($D_{\text{avg}}$ and $D^{EW}_{\text{avg}}$) indicate that the LLM can generally identify and assign appropriate scores for quality issues in the NLG task, regardless of the specific type of perturbation. The average discernment scores are useful for getting a broad understanding of an LLM's overall performance as an NLG evaluator. On the other hand, the minimum discernment scores $D_\text{min}$ and $D^{EW}_\text{min}$ assess the LLM's performance in the most challenging scenarios, where it may struggle to identify certain types of quality issues. These scores represent the lowest discernment score achieved by the LLM across all perturbation methods, indicating its weakest performance. The minimum discernment scores are crucial for understanding the limitations and potential failure modes of an LLM as an NLG evaluator, even if its overall average performance is acceptable. 

\section{Benchmarking LLM Discernment}

We evaluate five series of LLMs with varying parameter sizes: the GPT-series~\cite{wang2023chatgpt, openai2023gpt4turbo}, which includes GPT3.5-Turbo and GPT4-Turbo; the Llama3-series~\cite{meta2024llama3}; the Vicuna1.5 series~\cite{vicuna2023}; Mistral-7B~\cite{jiang2023mistral}; and the Qwen-series~\cite{qwen}. 

The LLMs are evaluated across four NLG tasks using six re-established public datasets: for \textbf{Summarization}, we use SummEval~\cite{fabbri2021summeval} (news articles) and SumPubMed~\cite{sumpubmed} (scientific articles); for \textbf{Story Completion}, we select data from Story Cloze Test dataset~\cite{storycloze}; for \textbf{Question Answering}, we utilize the data and modify the quality metric based on the Answer Equivalence dataset~\cite{answereq}; and for \textbf{Translation}, we leverage WMT-22 German-to-English and Chinese-to-English general (news) translation subsets~\cite{wmt22}. To ensure comparability, we select $N=100$ datapoints from each dataset. The quality metrics and perturbation methods are detailed in Table~\ref{table1}. 


We present our DHP benchmarking results in Figure~\ref{fig4}. By examining the discernment scores achieved by these models, we can gain insights into their competence as NLG evaluators.

\subsection{Overall Assessment} Most LLMs that we have evaluated demonstrate the ability to discern quality issues, as indicated by most $D_\text{avg}$ and $D^{EW}_\text{avg}$ scores exceeding 1. This suggests they can comprehend most evaluation metrics and detect varying quality in NLG tasks. However, an exception is noted in the WMT22 Chinese-to-English Translation dataset in Figure~\ref{fig4}~(f), where Vicuna1.5-7B and Qwen1.5-7B fail to achieve favorable average discernment scores, possibly due to their weaker multi-lingual capabilities.

Overall, for NLG evaluation, we recommend the GPT series, especially GPT4-Turbo, which demonstrates superior stability and the highest discernment across nearly all tasks. Among open-source models, Vicuna1.5-13B and Llama3-70B are commendable, achieving good average discernment scores and with most $D_{\text{min}}$ and $D^{EW}_{\text{min}}$ above 1.

\subsection{Other Observations}
\noindent\textbf{Trends regarding the size of LLMs:} In general, larger models within one series generally show better discernment. However, there are notable inconsistencies. For example, Qwen1.5-4B unexpectedly outperforms Qwen-7B in translation tasks in Figure~\ref{fig4}~(e, f), and Qwen-72B displays variable performance in the Question Answering task in Figure~\ref{fig4}~(d), suggesting that not all larger models uniformly perform better across all types of tasks.

\noindent\textbf{Limitations of Smaller LLMs:} In more challenging scenarios, represented by $D_\text{min}$ and $D^{EW}_\text{min}$, smaller-sized LLMs underperform. Models with fewer than 8B parameters show significantly lower $D_\text{min}$ and $D^{EW}_\text{min}$, particularly in summarization and translation tasks in Figure~\ref{fig4}~(a, b, e, f). Among these smaller models, Llama3-8B and Mistral-7B are relatively competitive with higher average scores but still register very low scores in the summarization tasks. This suggests that smaller models may become unstable and unreliable evaluators in some complex NLG evaluation scenarios.

\noindent\textbf{Metric Misunderstanding Phenomenon:} Differences between discernment scores with and without expert weights ($D$ and $D^{EW}$) are also notable. While most LLMs display consistent $D$ and $D^{EW}$ scores, Llama3-8B's performance in translation tasks in Figure~\ref{fig4}~(e, f) shows a significant discrepancy, with $D^{EW}_\text{min}$ values being substantially lower than $D_\text{min}$ and even dropping below 1. This indicates the model's misunderstanding in metrics while identifying quality issues.

\noindent\textbf{Variations in Task Performance:} Among the six datasets, LLMs perform best in the Story Cloze Test in Figure~\ref{fig4}~(c), achieving higher and more stable scores. However, the SumPubMed dataset presented in Figure~\ref{fig4}~(b) proves the most challenging; all models except GPT4-Turbo score below 1 in $D_\text{min}$ and $D^{EW}_\text{min}$ because of the dataset's complex scientific terminology and content. Models lacking sufficient prior knowledge struggle to identify subtle quality issues in such specialized content. Therefore, we encourage the community to test LLM discernment scores for their specific NLG tasks prior to conducting evaluations, ensuring the selected models are competent evaluators.

\section{Related Work}


Recent advancements highlight the significant potential of utilizing LLMs as evaluators for a variety of NLP tasks. Extensive empirical evidence supports this viewpoint, as demonstrated by studies~\cite{liu2023g, chiang2023can, hu2024llm, desmond2024evalullm, wang2023chatgpt}, which assert that the evaluation behaviors of pretrained LLM-based evaluators are well-aligned with those of human preference~\cite{liu2023calibrating}. Liusie et al.\cite{liusie2024llm} further show that comparative assessments using LLM evaluators outperform prompt-based techniques, though they identify potential positional biases and propose corresponding solutions. Despite the great assessment performance of a single LLM, advanced studies involve multi-LLM agents~\cite{chan2023chateval, zhang2023wider, li2023prd} or human experts~\cite{gao2024llm, li2024leveraging} to further increase the judging capability.

While the application of LLMs as judges is a burgeoning area of research, it is imperative to assess their reliability and effectiveness in evaluative roles. To this end, several benchmarks have been recently proposed to evaluate LLMs as judges. For example, JudgeBench ~\cite{tan2024judgebench} is designed to assess LLM-based judges on challenging response pairs spanning knowledge, reasoning, math, and coding. Additionally, LLM-judge-eval ~\cite{wei2024systematic} evaluates tasks such as summarization and alignment, incorporating metrics like flipping noise and length bias.

However, despite the progress in LLMs as judges, several challenges persist. First, human involvement remains a crucial factor in both evaluation and alignment, which raises concerns about the extent to which human biases influence LLM-based evaluations. Second, human evaluators themselves are inherently biased, meaning that even if an LLM aligns well with human preferences, it does not necessarily guarantee fairness or accuracy. Additionally, LLMs may misinterpret NLG evaluation metrics~\cite{hu2024llm}, making simple alignment scores unreliable. To overcome these challenges, our work focuses on developing automated and comprehensive methodologies to test the reliability of LLM-based evaluations.

\section{Conclusion}

We introduce the DHP benchmark to assess the discernment capabilities of LLMs as evaluators across various NLG tasks. Our approach not only provides benchmarking results for LLMs but also establishes a robust framework to evaluate how effectively LLMs can identify quality issues, thus serving as competent NLG evaluators. While most models generally perform well, their performance is significantly influenced by factors such as model size, task type, and dataset complexity. By identifying specific weaknesses of LLMs in evaluating NLG tasks, this benchmark aids researchers in enhancing ``LLM-as-a-judge'' methodologies and improving overall LLM performance.

\section{Limitations}
While our DHP benchmark provides a systematic and scalable way to assess LLMs' ability to detect targeted quality issues, it does not offer a complete picture of how these models perform on every aspect of NLG evaluation. First, the discernment scores are generated on a dataset-by-dataset basis, so a truly comprehensive assessment of LLMs across different NLG tasks remains an open challenge. Next, although our framework is designed to reduce reliance on human annotations, it does not fully replace the depth and contextual insight that human evaluations provide. Our perturbation-driven approach highlights particular types of errors rather than capturing the broad spectrum of real-world NLG complexities. Consequently, DHP is best viewed as a complementary tool, and further work is needed to expand its scope to more diverse tasks, languages, and cultural settings, as well as to integrate human judgment for a more holistic evaluation of LLMs.
\bibliography{custom}

\clearpage
\appendix
\onecolumn

\section{NLG Tasks and Metrics}
\subsection{Summarization}

We utilize the SummEval~\cite{fabbri2021summeval} (MIT license) and SumPubMed~\cite{sumpubmed} datasets (MIT license) for our summarization tasks. The SummEval dataset comprises 100 news articles, each accompanied by multiple reference and generated summaries. For our analysis, we exclusively use the reference summaries, selecting the one with the highest number of sentences from each article to facilitate perturbation. The SumPubMed dataset contains 32,000 long scientific articles along with their abstracts serving as reference summaries. We only use the "BACKGROUND" sections of these articles and summaries. We randomly select 100 pairs of articles and their corresponding summaries.

For the evaluation of summarization performance, we adhere to the metrics defined by SummEval~\cite{fabbri2021summeval}, specifically focusing on Coherence, Consistency, Fluency, and Relevance.

\subsection{Story Completion}

In this story completion task, we utilize the public Story Cloze Test dataset~\cite{storycloze}, which comprises four-sentence stories each paired with a reference and wrong ending. We select 100 datapoints at random from the validation set for our analysis.

Given the absence of explicitly defined quality metrics for the dataset, we adapt metrics from summarization tasks—Coherence, Consistency, and Fluency. Coherence evaluates the story's overall structure and narrative flow. Consistency measures how well the ending maintains the established tone, setting, character development, and narrative style of the story. Fluency focuses on the linguistic and stylistic quality of the story's conclusion.

\subsection{Question Answering}

For the question answering task, we employ the Answer Equivalence dataset~\cite{answereq} (Apache-2.0 license), which is a modified version of the SQuAD dataset~\cite{rajpurkar2016squad}. We specifically select reference answers that exceed 150 characters to facilitate perturbation. From this filtered set, we randomly choose 100 question-answer pairs.

We adapt the original rating tasks of the dataset into a single metric: Answer Quality. This metric assesses whether the answer provides a comprehensive and accurate response to the question, effectively capturing the essence of the content discussed in the paragraph.

\subsection{Translation}
We utilize two subsets from the WMT-22 general (news) translation dataset: German-to-English and Chinese-to-English sets which are freely available for research purposes. For our analysis, we select the test sets with reference translations, ensuring each translation exceeds 300 characters in length. We randomly choose 100 datapoints from each subset for evaluation.

In assessing translation tasks, we adopt two principal metrics from the Multidimensional Quality Metrics (MQM) framework~\cite{mqm}: Accuracy and Fluency. Accuracy measures how closely the translation mirrors the source text, focusing on the absence of additions, omissions, or mistranslations. Fluency evaluates the translation's compliance with the linguistic norms of the target language, specifically examining spelling, grammar, and consistency.

\begin{table*}[!t]
\small
\caption{Summary of hierarchical perturbation methods applied to different NLG tasks, detailing the types of perturbations and their respective implementations based on character (C), word (W), and sentence-level (S) modification with rule-based (R) or LLM-based (L) approaches.}
\centering
\scalebox{0.85}{
\begin{tabular}{llll}
\toprule
\textbf{Task}                                                                          & \textbf{Avg NLTK Statistics}                                                                                                                                                                                                & \textbf{Perturbation}                    & \textbf{Description}                                                                                                                                                                                                                                                 \\ \midrule
\multirow{6}{*}{Summarization}                                                & \multirow{6}{*}{\begin{tabular}[c]{@{}l@{}}SummEval:\\ 340.4 Characters\\ 58.3 Words\\ 4.0 Sentences \\ \\ SumPubMed\\ 803.5 Characters\\ 114.9 Words\\ 5.5 Sentences\end{tabular}}                           & (C, R) Random Deletions         & \begin{tabular}[c]{@{}l@{}}Delete k alphanumeric characters randomly.\\ \\ SummEval: k=10 for Minor, k=50 for Major;\\ SumPubMed: k=20 for Minor, k=100 for Major.\end{tabular}                                                                             \\ \cmidrule{3-4} 
    &                                                                                                                                                                                                               & (C, R) Random Typos             & \begin{tabular}[c]{@{}l@{}}Add k random typographical errors with "typo" package.\\ \\ SummEval: k=10 for Minor, k=50 for Major;\\ SumPubMed: k=20 for Minor, k=100 for Major.\end{tabular}                                                              \\ \cmidrule{3-4} 
    &                                                                                                                                                                                                               & (W, L) Fictional Named Entities & \begin{tabular}[c]{@{}l@{}}Substitute one ore more named entities with in the \\ summary (e.g., names, locations, specific numbers, \\ technical terms, etc.) with fictional counterparts.\end{tabular}                                                     \\ \cmidrule{3-4} 
    &                                                                                                                                                                                                               & (W, L) Grammatical Errors       & \begin{tabular}[c]{@{}l@{}}Modify the summary for creating two or more \\ grammatical errors, such as subject-verb \\ disagreement, noun-pronoun disagreement, \\ incorrect verb tense, misuse of preposition, \\ and sentence fragment, etc.\end{tabular}  \\ \cmidrule{3-4} 
    &                                                                                                                                                                                                               & (S, R) Reordering               & \begin{tabular}[c]{@{}l@{}}Random shuffle k sentences in the summary.\\ \\ k=2 for Minor, k=all for Major.\end{tabular}                                                                                                                                     \\ \cmidrule{3-4} 
    &                                                                                                                                                                                                               & (S, L) Rewriting and Insertion  & \begin{tabular}[c]{@{}l@{}}Select one or more sentences from the summary, \\ then rephrase them and insert the rewritten \\ versions immediately after the original sentences.\end{tabular}                                                                 \\ \midrule
\multirow{6}{*}{\begin{tabular}[c]{@{}l@{}}Story\\ Completion\end{tabular}}   & \multirow{6}{*}{\begin{tabular}[c]{@{}l@{}}Story Cloze Test:\\ 38.7 Characters\\ 7.4 Words\\ 1.0 Sentences\end{tabular}}                                                                                      & (C, R) Random Deletions         & Delete 5 alphanumeric characters randomly.                                                                                                                                                                                                                  \\ \cmidrule{3-4} 
    &                                                                                                                                                                                                               & (C, R) Random Typos             & \begin{tabular}[c]{@{}l@{}}Add 5 random typographical errors with "typo" package.\end{tabular}                                                                                                                                                            \\ \cmidrule{3-4} 
    &                                                                                                                                                                                                               & (W, L) Fictional Named Entities & \begin{tabular}[c]{@{}l@{}}Substitute one critical named entities within the ending\\ sentence (e.g., a name, a location, a specific number, etc.) \\ with a fictional counterpart.\end{tabular}                                                            \\ \cmidrule{3-4} 
    &                                                                                                                                                                                                               & (W, L) Grammatical Errors       & \begin{tabular}[c]{@{}l@{}}Modify the ending for creating one grammatical error, \\ such as subject-verb disagreement, noun-pronoun \\ disagreement, incorrect verb tense, misuse of preposition, \\ and sentence fragment, etc.\end{tabular}               \\ \cmidrule{3-4} 
    &                                                                                                                                                                                                               & (S, R) Random Ending Sentence   & Replace the ending with a random one from another story.                                                                                                                                                                                                    \\ \cmidrule{3-4} 
    &                                                                                                                                                                                                               & (S, R) Wrong Ending Sentence    & Replace the ending with the wrong ending of the dataset.                                                                                                                                                                                                    \\ \midrule
\multirow{5}{*}{\begin{tabular}[c]{@{}l@{}}Question\\ Answering\end{tabular}} & \multirow{5}{*}{\begin{tabular}[c]{@{}l@{}}Answer Equivalence:\\ 156.2 Characters\\ 23.9 Words\\ 1.0 Sentences\end{tabular}}                                                                                  & (C, R) Random Deletions         & \begin{tabular}[c]{@{}l@{}}Delete k alphanumeric characters randomly.\\ \\ k=5 for Minor, k=25 for Major.\end{tabular}                                                                                                                                      \\ \cmidrule{3-4} 
    &                                                                                                                                                                                                               & (C, R) Random Typos             & \begin{tabular}[c]{@{}l@{}}Add k random typographical errors with "typo" package.\\ \\ k=5 for Minor, k=25 for Major.\end{tabular}                                                                                                                \\ \cmidrule{3-4} 
    &                                                                                                                                                                                                               & (W, L) Fictional Named Entities & \begin{tabular}[c]{@{}l@{}}Substitute one or more critical named entities within \\ the answer (e.g., names, locations, specific numbers, \\ technical terms, etc.) with fictional counterparts.\end{tabular}                                               \\ \cmidrule{3-4} 
    &                                                                                                                                                                                                               & (W, L) Grammatical Errors       & \begin{tabular}[c]{@{}l@{}}Modify the answer for creating one or more grammatical \\ errors, such as subject-verb disagreement, noun-pronoun \\ disagreement, incorrect verb tense, misuse of preposition, \\ and sentence fragment, etc.\end{tabular}      \\ \cmidrule{3-4} 
    &                                                                                                                                                                                                               & (S, R) Random Answer            & Replace the answer with a random one to another question.                                                                                                                                                                                                   \\ \midrule
\multirow{5}{*}{Translation}                                                  & \multirow{5}{*}{\begin{tabular}[c]{@{}l@{}}WMT-22 \\ German-to-English:\\ 436.8 Characters\\ 71.0 Words\\ 3.8 Sentences\\ \\ WMT-22 \\Chinese-to-English:\\ 434.1 Characters\\ 66.4 Words\\ 1.1 Sentences\end{tabular}} & (C, R) Random Deletions         & \begin{tabular}[c]{@{}l@{}}Delete k alphanumeric characters randomly.\\ \\ k=10 for Minor, k=50 for Major.\end{tabular}                                                                                                                                     \\ \cmidrule{3-4} 
    &                                                                                                                                                                                                               & (C, R) Random Typos             & \begin{tabular}[c]{@{}l@{}}Add k random typographical errors with "typo" package.\\ \\ k=10 for Minor, k=50 for Major.\end{tabular}                                                                                                               \\ \cmidrule{3-4} 
    &                                                                                                                                                                                                               & (W, R) Random Deletions         & \begin{tabular}[c]{@{}l@{}}Delete k continuous words in the translation randomly.\\ \\ k=5 for Minor, k=25 for Major.\end{tabular}                                                                                                                          \\ \cmidrule{3-4} 
    &                                                                                                                                                                                                               & (W, L) Fictional Named Entities & \begin{tabular}[c]{@{}l@{}}Substitute one or more critical named entities within \\ the translation (e.g., names, locations, specific numbers, \\ technical terms, etc.) with fictional counterparts.\end{tabular}                                          \\ \cmidrule{3-4} 
    &                                                                                                                                                                                                               & (W, L) Grammatical Errors       & \begin{tabular}[c]{@{}l@{}}Modify the translation for creating two or more \\ grammatical errors, such as subject-verb disagreement, \\ noun-pronoun disagreement, incorrect verb tense, misuse \\ of preposition, and sentence fragment, etc.\end{tabular} \\ \bottomrule
\end{tabular}}
\label{table-perturb}

\end{table*}

\clearpage
\section{Hierarchical Perturbation}

The specifics of the hierarchical perturbations are detailed in Table~\ref{table-perturb}. We perform these perturbations based on character, word, and sentence-level statistical data of the texts, which are presented in Table~\ref{table-perturb}. Our rule-based perturbations include simple text deletions, typographical errors using existing software tools, reordering of sentences, and the incorporation of random or incorrect sentences from other data.

For LLM-based perturbations, we employ GPT4-Turbo, modifying the reference text via Auto-CoT~\cite{zhang2022automatic} prompts to generate the detailed procedural perturbation steps. Below, we provide an example of how the ``Minor Fictional Named Entities'' perturbation is applied to the summarization tasks:

Minor Fictional Named Entities Perturbation Prompt:

{\itshape
You will be given one summary written for an article. 
Your task is to adjust the summary by implementing a specific change.

Please make sure you read and understand these instructions carefully.

Adjustment:
Please substitute only one critical named entity within the summary (e.g., a name, a location, a specific number, a technical term, etc.) with a fictional counterpart.

Adjustment Steps:

1. Identify the critical named entity within the summary. This could be a person's name, a location, a specific number, or any other specific detail that is crucial to the summary.

2. Create a fictional counterpart for the identified entity. This could be a fictional name, a fictional location, a fictional number, a fictional technical term etc. Make sure that the fictional counterpart is appropriate and fits within the context of the summary.

3. Replace the identified entity with its fictional counterpart in the summary. Ensure that the replacement is grammatically correct and maintains the overall meaning and flow of the summary.

4. Review the adjusted summary to ensure that it still makes sense and conveys the main points of the article, despite the change in one critical named entity.

Summary:

SUMMARY\_HERE

Revised Summary:
}

\begin{figure*}[!t]
    \centering
    \includegraphics[width=0.95\textwidth]{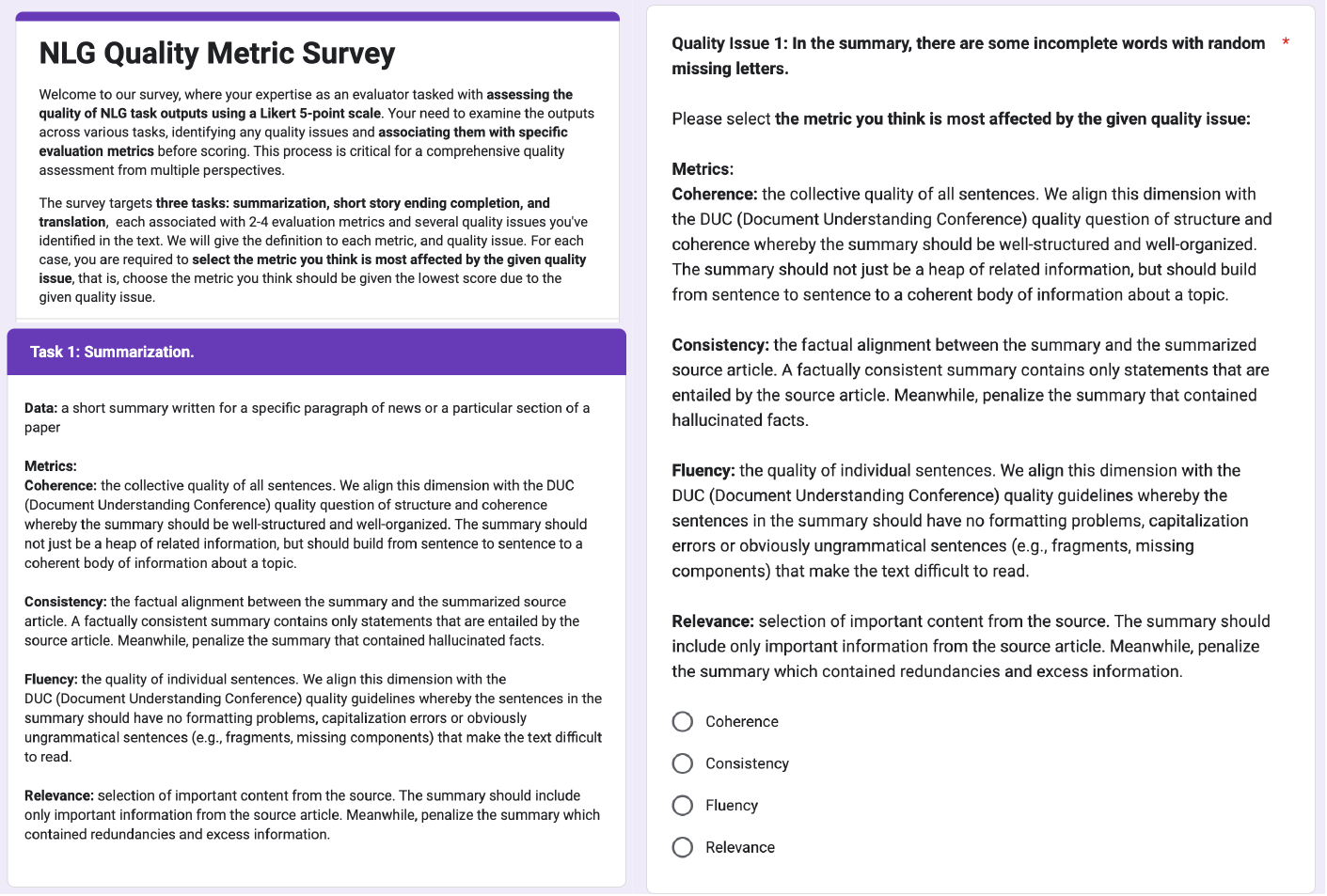}
    \caption{User interface of the expert weight survey conducted to determine the impact of various quality issues on NLG task metrics.}
    \label{apdx_fig1}
\end{figure*}

\begin{figure*}[!t]
    \centering
    \includegraphics[width=\textwidth]{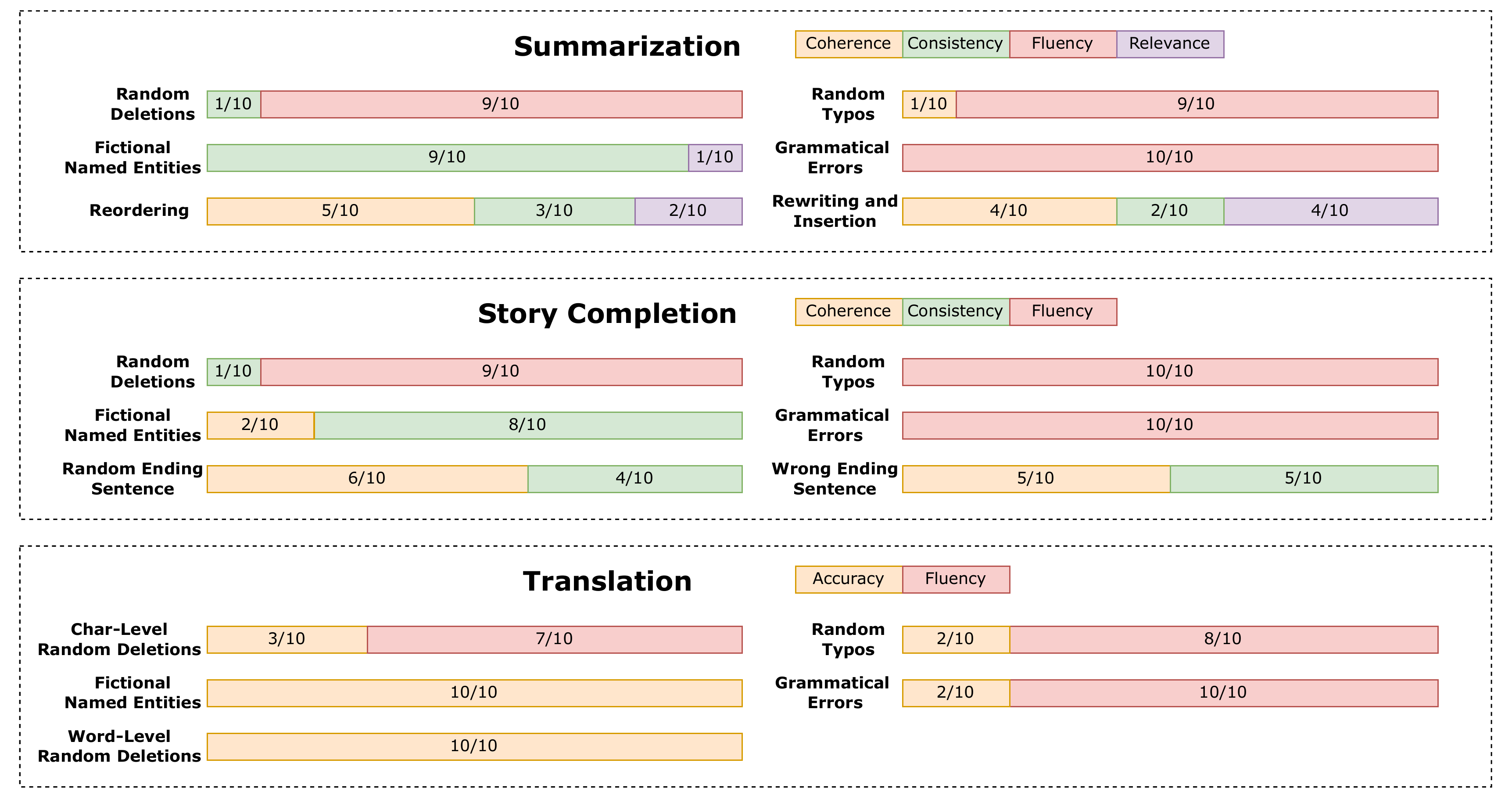}
    \caption{Graphical representation of the expert weights for each NLG task.}
    \label{apdx_fig2}
\end{figure*}

\section{Expert Weights}
We invite 10 volunteer experts with extensive backgrounds in NLP/NLG research to complete an expert weight survey. The interface of this survey is displayed in Figure~\ref{apdx_fig1}, which includes the survey instructions, definitions of the tasks and metrics, data types, and descriptions of quality issues associated with the perturbation methods. The experts are asked to select the metric they believe is most impacted by each quality issue presented. We then utilize their responses as weights for combining the $p$-values. The results of these expert evaluations are detailed in Figure~\ref{apdx_fig2}.

\begin{table*}[t]
\caption{Overview of large language models (LLMs) assessed in the DHP benchmark, specifying model versions and sources.}
\centering
\footnotesize
\scalebox{0.9}{
\begin{tabular}{lll}
\toprule
Model         & Version                   & Source                                              \\ \midrule
GPT3.5-Turbo  & gpt-3.5-turbo-0125        & \url{platform.openai.com/docs/models}                     \\
GPT4-Turbo    & gpt-4-1106-preview        & \url{platform.openai.com/docs/models}                     \\ \midrule
Llama3-8B     & Meta-Llama-3-8B-Instruct  & \url{huggingface.co/meta-llama/Meta-Llama-3-8B-Instruct}  \\
Llama3-70B    & Meta-Llama-3-70B-Instruct & \url{huggingface.co/meta-llama/Meta-Llama-3-70B-Instruct} \\ \midrule
Vicuna1.5-7B  & vicuna-7b-v1.5-16k        & \url{huggingface.co/lmsys/vicuna-7b-v1.5-16k}             \\
Vicuna1.5-13B & vicuna-13b-v1.5-16k       & \url{huggingface.co/lmsys/vicuna-13b-v1.5-16k}            \\ \midrule
Mistral-7B    & Mistral-7B-Instruct-v0.2  & \url{huggingface.co/mistralai/Mistral-7B-Instruct-v0.2}   \\ \midrule
Qwen1.5-4B    & Qwen1.5-4B-Chat           & \url{huggingface.co/Qwen/Qwen1.5-4B-Chat}                 \\
Qwen1.5-7B    & Qwen1.5-7B-Chat           & \url{huggingface.co/Qwen/Qwen1.5-7B-Chat}                 \\
Qwen1.5-14B   & Qwen1.5-14B-Chat          & \url{huggingface.co/Qwen/Qwen1.5-14B-Chat}                \\
Qwen1.5-32B   & Qwen1.5-32B-Chat          & \url{huggingface.co/Qwen/Qwen1.5-32B-Chat}                \\
Qwen1.5-72B   & Qwen1.5-72B-Chat          & \url{huggingface.co/Qwen/Qwen1.5-72B-Chat}                \\ \bottomrule
\end{tabular}}
\label{table_models}
\end{table*}

\clearpage
\section{LLM Evaluation}
\vspace{-0.2cm}
We evaluate five series of large language models (LLMs), details of which are provided in Table~\ref{table_models}. Due to the extensive length of text data from the SumPubMed dataset~\cite{sumpubmed}, which can exceed the 4K context window, we evaluate the models capable of processing long texts ($\geq$ 8K tokens). The GPT series is operated using the \href{https://platform.openai.com/docs/overview}{OpenAI API}, and the open-source LLMs are executed on a server with 8 Nvidia A100 GPUs. We set the temperature parameters to 0 and maintain the default values for the top\_p parameters. Throughout the evaluation process, each model score 5 times on each metric to calculate a final average score. We use the \href{https://docs.scipy.org/doc/scipy/reference/generated/scipy.stats.wilcoxon.html}{\textit{scipy.stats.wilcoxon}} to conduct the Wilcoxon Signed-Rank Test.

\section{Evaluation Prompts}
\vspace{-0.2cm}
We follow the guidelines of G-Eval~\cite{liu2023g} and utilize the Auto-CoT method~\cite{zhang2022automatic} to construct our evaluation prompts. Below is an example of the prompt used for assessing the Coherence metric in summarization tasks:

{\itshape
You will be given a summary written for an article. Your task is to rate the summary on one metric. Please make sure you read and understand these instructions carefully. Please keep this document open while reviewing, and refer to it as needed.

Evaluation Criterion: Coherence (1-5) - the collective quality of all sentences. We align this dimension with the DUC quality question of structure and coherence whereby the summary should be well-structured and well-organized. The summary should not just be a heap of related information, but should build from sentence to sentence to a coherent body of information about a topic. 

Evaluation Steps:

1. Read the Summary Thoroughly: Before diving into the evaluation, ensure that you have a clear understanding of the entire summary. Reading it more than once might be necessary.

2. Identify the Central Topic: A coherent summary will have a clear central topic or theme. Identify this topic and see if the subsequent information revolves around it.

3. Check for Logical Flow: Review the summary for logical sequencing. Sentences should follow one another in a way that makes sense and allows the reader to easily follow the progression of information.

4. Look for Transitional Elements: Coherent summaries often have clear transitions between sentences or ideas. This could be in the form of transitional words, phrases, or connecting ideas that tie one sentence to the next.

5. Identify Redundancies: Check if the same information is repeated in different sentences. Redundancies can disrupt the flow and coherence of a summary.

6. Note Any Gaps or Jumps: If there are sudden jumps in topics or if crucial information seems to be missing, this can harm the coherence of the summary. A well-organized summary should present a holistic view of the topic without leaving the reader with questions.

7. Assess Clarity: Even if the content is technically accurate, if it's written in a convoluted or unclear manner, it can disrupt coherence. The sentences should be clear and easily understandable.

8. Consider the Conclusion: A coherent summary often wraps up or comes to a conclusion that ties the presented information together. It doesn’t necessarily need a formal conclusion, but the end should feel natural and not abrupt.

9. Rate the Summary: Based on the above steps, assign a score between 1-5 for coherence. 
    - 1: Very incoherent. The summary lacks structure, has sudden jumps, and is difficult to follow.
    - 2: Somewhat incoherent. The summary has some semblance of structure, but has significant flaws in flow and organization.
    - 3: Neutral. The summary is decently organized, with minor issues in flow and structure. 
    - 4: Mostly coherent. The summary is well-structured with very few minor coherence issues.
    - 5: Highly coherent. The summary is excellently organized, flows seamlessly, and builds information logically from start to end.

Source Article: 

ARTICLE\_HERE

Summary: 

SUMMARY\_HERE

Evaluation Score (please don't give any feedback, just give a score ONLY) - Coherence:
}

\end{document}